\documentclass[final]{cvpr}

\usepackage{times}
\usepackage{epsfig}
\usepackage{graphicx}
\usepackage{amsmath}
\usepackage{amssymb}
\usepackage{caption}
\usepackage{mathtools}

\usepackage[pagebackref=true,breaklinks=true,colorlinks,bookmarks=false]{hyperref}

\newcommand{\boldstart}[1]{\noindent\textbf{#1}}

\begin{document}

%%%%%%%%% TITLE
\vspace{-1cm}
\title{ANR: Articulated Neural Rendering for Virtual Avatars}

\author{\vspace{-1.25em}%
Amit Raj ${}^{1}$\quad 
Julian Tanke${}^{2}$\quad 
James Hays${}^1$ \\ \quad \\%
Minh Vo${}^3$\quad %
Carsten Stoll${}^4$\quad %
Christoph Lassner${}^3$%
\\ \quad \\ 
${}^1$Georgia Tech\quad%
${}^2$University of Bonn\quad%
${}^3$Facebook Reality Labs
${}^4$Epic Games
}

\twocolumn[{%
\renewcommand\twocolumn[1][]{#1}%
\vspace{-1em}
\maketitle
\vspace{-1em}
\begin{center}
    \centering
    \vspace{-0.3in}
    \includegraphics[width=\textwidth]{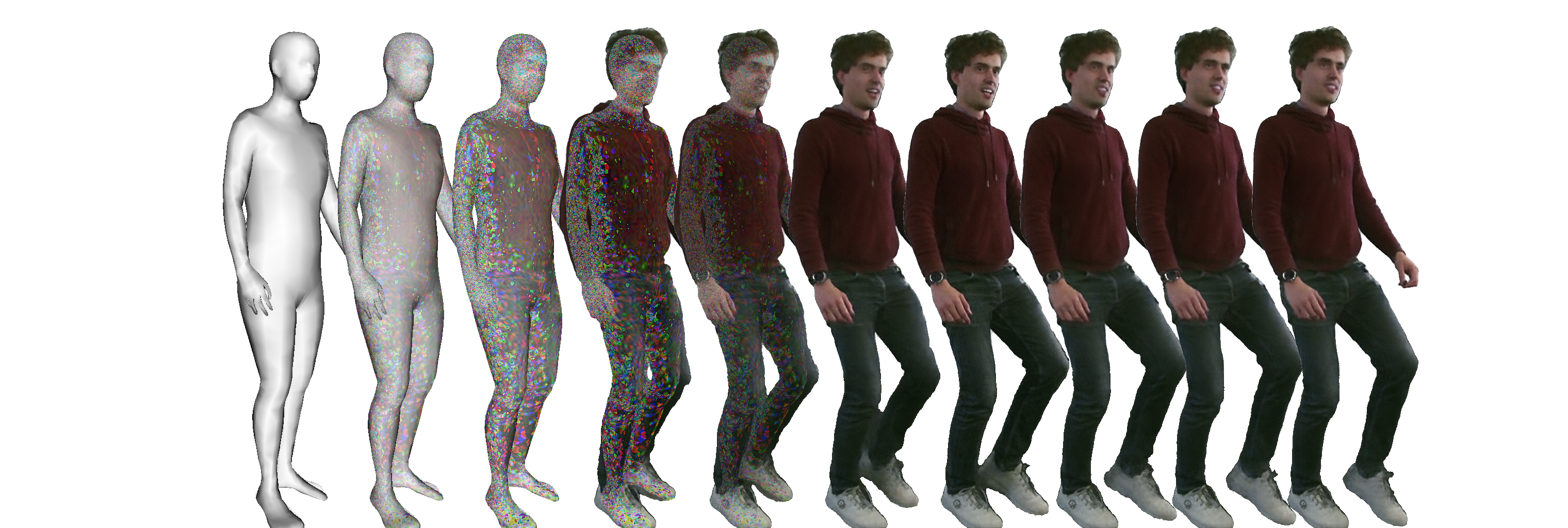}
    \captionof{figure}{We present Articulated Neural Rendering (ANR), a rendering framework capable of producing highly realistic avatars. Similar to Deferred Neural Rendering (DNR)~\cite{thies2019deferred}, ANR uses neural networks to convert a latent texture on a coarse mesh (left) into an RGB image (right). Unlike DNR, which is ineffective when the mesh geometry is inaccurate or deforms during motion, ANR explicitly accounts for such geometric misalignment and pose-dependent deformation.
    }
    \label{fig:teaser}
\end{center}%
}]
\begin{abstract}
\vspace{-1cm}
    The combination of traditional rendering with neural networks in Deferred Neural Rendering (DNR)~\cite{thies2019deferred} provides a compelling balance between computational complexity and realism of the resulting images. Using skinned meshes for rendering articulating objects is a natural extension for the DNR framework and would open it up to a plethora of applications. However, in this case the neural shading step must account for deformations that are possibly not captured in the mesh, as well as alignment inaccuracies and dynamics---which can confound the DNR pipeline. We present Articulated Neural Rendering (ANR), a novel framework based on DNR which explicitly addresses its limitations for virtual human avatars. We show the superiority of ANR not only with respect to DNR but also with methods specialized for avatar creation and animation. In two user studies, we observe a clear preference for our avatar model and we demonstrate state-of-the-art performance on   quantitative evaluation metrics. Perceptually, we observe better temporal stability, level of detail and plausibility. More results are available at our project page: \url{https://anr-avatars.github.io}.
\end{abstract}

\section{Introduction}

\begin{figure*}
    \centering
    \includegraphics[width=0.95\linewidth]{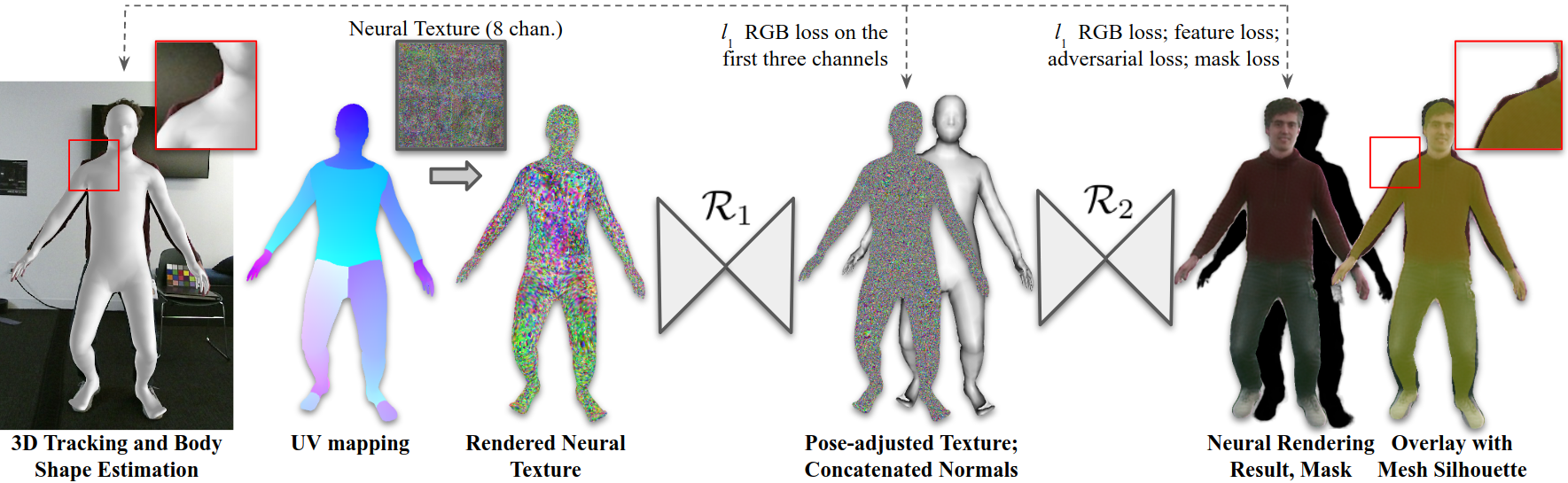}
    \vspace*{-0.3cm}
    \caption{Schematic overview of the proposed framework. Given a coarse, animated 3D body mesh, ANR produces a detailed avatar. Using rasterized IUV images of the mesh using a weak perspective projection, we render an 8 channel neural texture into image space. A first stage, $\mathcal{R}_1$, transforms the texture into another, refined latent representation, which we combine with the normal information. The second stage $\mathcal{R}_2$ uses this information to create an RGB rendering and a foreground mask. The rendering can extend beyond the coarse mesh, in this case we visualize only slight refinement for painting the shirt.}
    \label{fig:framework}
\end{figure*}

%Motivation.
Capturing realistic appearance is one of the important goals of computer vision. Progress in 3D rendering and neural networks has led to approaches with remarkable fidelity~\cite{lombardi2018deep,lombardi2019neural,mildenhall2020nerf,nagano2019deep}. These methods often use expensive and intricate capture setups which prevent easy digitization and transfer of the resulting models~ \cite{collet2015high,debevec2000acquiring,guo2019relightables}. The recent deferred neural rendering paradigm offers an exciting opportunity to work with inaccurate geometry and relatively simple neural shaders while capturing complex scenes with view-dependent effects realistically~\cite{aliev2019neural, martin2020gelato,thies2019deferred}. In a first step, the geometry is rasterized using a \textit{neural} latent texture which is then translated to an RGB image using a convolutional network. Both, the rendering network as well as the neural texture, are optimized to produce realistic results. 

% Problem.
Deferred neural rendering works particularly well for rigid objects. Its pipeline could be extended to deformable objects in a natural way: a \emph{skinned} mesh could be used for capturing the geometry. The rasterized neural texture from the posed mesh could then be translated to an RGB image. While this idea is conceptually simple, the neural network has to learn more complex deformation-dependent effects. Furthermore, the mesh used for rendering is usually not a perfect representation of the real geometry, leading to alignment problems. These problems are currently not taken into account~\cite{aliev2019neural,martin2020gelato,thies2019deferred}, which limits the application of DNR in scenarios with deformable objects.

% Our proposed solution.
In this paper, we present Articulated Neural Rendering~(ANR) to account for these problems. ANR systematically rebuilds DNR from the neural shading model architecture to the optimization scheme. We use ANR to tackle one of the most challenging problems for animation: virtual human avatars. Fig.~\ref{fig:teaser} shows an example of an avatar rendered using ANR.

% Approach overview.
Our method employs a simple statistical human body model fitted to a training video to capture the body shape statistics and 3D pose information for each frame~\cite{walsman2017dynamic}. This body model only represents the coarse body geometry without clothing and hair. Consequently, direct use of the DNR pipeline leads to unrealistic and blurry results. We use keyframes from the video to learn the \textit{static} appearance encoded in the neural texture, and use the other frames to learn the \textit{dynamic} pose-conditioned rendering of the appearance. Our keyframes-based training scheme enables the model to converge 5X faster and produces quantitatively better avatars than DNR. We simultaneously train ANR on multiple identities in a single model, leading to decoupling of the neural texture and the shading model. Owing to the consistent surface parameterization of the statistical body model, our model can leverage such semantic correspondences to modify and mix components from multiple neural textures, enabling virtual try-on by changing regions in the neural texture. While our model works solely in 2D, we experimentally validate that it can render near photorealistic and persistent 3D appearance of people with a very small network (161M parameters). In two user studies, we demonstrate that we not only outperform the DNR pipeline, but also several methods dedicated to creating virtual avatars~\cite{shysheya2019textured, wang2018vid2vid}. Perceptually, the presented method is temporally stable and captures fine appearance details.

Our contributions are threefold:
\begin{itemize}
\item We propose ANR, a novel neural rendering framework, to generate high-quality virtual avatars from coarse 3D shape and arbitrary skeletal motions. Key to ANR is to account for geometric misalignment of the coarse body mesh and pose-dependent deformation.
\item Using ANR, we present the first neural avatar model that can capture and render multiple identities with only one set of network parameters in addition to an identity specific neural texture map.
\item The explicit texture space makes it possible to edit or mix identities, which is novel in the context of neural rendering for avatars.
\end{itemize}

For higher quality results, fine-tuning the model on a specific identity is an option. This makes the resulting avatars directly applicable in use cases where the range of motion is known or can be estimated well, for example for virtual assistants or game characters. 

\section{Related Work}

Among many methods to create and render articulated models, a majority of them follow the classical pipeline of acquiring an accurate 4D geometry reconstruction with detailed textures painted on this geometry. Using machine learning, several recent methods have set out to perform inference mostly in 2D space, only using rough or no 3D guidance. We will discuss several frameworks from both of these schools as well as some hybrid methods, which are closest to our approach.

\boldstart{Inference in 3D Space:} The Relightables~\cite{guo2019relightables} propose a system that captures accurate geometry and texture using a controlled light stage. This allows for relightable rendering of the captured identity in different environments. Lombardi et al.~\cite{lombardi2018deep} use a multi camera setup to determine the average texture and deformations on a base mesh and use a neural network to generate view specific texture to render high fidelity images from different viewpoints. Using a similar system, Brualla et al.~\cite{martin2018lookingood} train a network to perform completion and super resolution of the rendered 3D model. In a single-view regime, Alldiek et al.~\cite{Alldieck_2019_CVPR,Alldieck_2018_CVPR} generate avatars by learning to regress accurate geometry and texture using purely synthetic data. Zhi et al.~\cite{zhi2020texmesh} estimate personalized avatar with fine geometry and texture by finetuning on the test video using self-supervised losses. DeepCap~\cite{deepcap} captures accurate geometry from monocular video by predicting a parameterized human configuration and deformation model. Our approach also uses monocular video to capture a digital avatar. %However, our avatar is coarsely parameterized by a low-dimensional statistical mesh and we must compensate for geometric inaccuracies that are not captured by this model (such as clothing and hairs) while rendering the model from a particular viewpoint.
However, instead of deforming and refining the avatar geometry parameterized by a statistical model, we advocate for avatars with high-capacity texture that compensates for such geometric inaccuracies (such as clothing and hairs) for arbitrary body pose and view-point rendering.

\boldstart{Inference in 2D Space:}
Meanwhile, specific architectures are designed for motion re-targeting, novel view synthesis and identity transfer, which primarily use only pixel and pose information~\cite{balakrishnan2018synthesizing,chan2019everybody,ma2018disentangled,si2018multistage,zhao2018multi}.
Neverova et al.~\cite{neverova2018dense} use DensePose for novel viewpoint synthesis, which is limited by the DensePose body coverage and accuracy. \cite{pandey2019volumetric} propose a semi-parametric approach which uses a combination of previously captured RGB(D) images and neural rendering to infer novel views in an approach similar to image based rendering.
The works by~\cite{ma2017pose,lassner2017generative} focus on pose conditioned image generation of people, but with lower resolution. Grigorev et al.~\cite{grigorev2018coordinate} solve the novel view synthesis problem by formulating it as texture inpainting in DensePose UV space. SwapNet~\cite{raj2018swapnet} learns to transfer clothing information by disentangling the notion of pose and clothing without being identity specific. Human appearance transfer~\cite{zanfir2018human} learns to generate novel views and transfer identities by performing human parsing and 3D shape and pose fitting. We generate 3D textured avatars, enabling all these tasks with no additional guiding signals or changes in a single framework.

\boldstart{Hybrid Approaches:} The DNR framework~\cite{thies2019deferred} uses a mostly rigid mesh and a neural texture to translate the rendering result into an image. We detail this approach in Sec.~\ref{sec:dnr} and reformulate it to account for handling fully articulated objects. Textured neural avatars~\cite{shysheya2019textured} present a framework to learn neural avatars in an end-to-end manner from multiview data. Unlike this work, we leverage the reconstructed geometry instead of noisy DensePose correspondences to generate the UV coordinates for every pixel, enabling us to maintain better texture consistency across viewpoints. Our work is also related to the Liquid Warping Gan~\cite{lwb2019} which performs appearance transfer and motion retargeting in a single network. However, our framework provides explicit access to the learned texture allowing for fine grained edits of appearance. Additionally, our framework uses a lower number of parameters, and thus can be trained at a higher resolution. Neural rendering and reenactment~\cite{Liu2018Neural} trains a network to translate from 3D pose to image. However, their framework involves capturing a rigged template mesh for every individual and requiring additional depth and body part information. Recently, implicit representations with impressive geometry reconstruction of clothed humans reconstruction from a single image are becoming popular~\cite{huang2020arch,saito2020pifuhd}. 
We distinguish ourselves from these methods by generating clothing and body deformations in the rendering stage while using a simple parametric body model to fit the body pose and shape.

\section{Approach}

\begin{figure*}[t!]
    \centering
    \includegraphics[width=\linewidth]{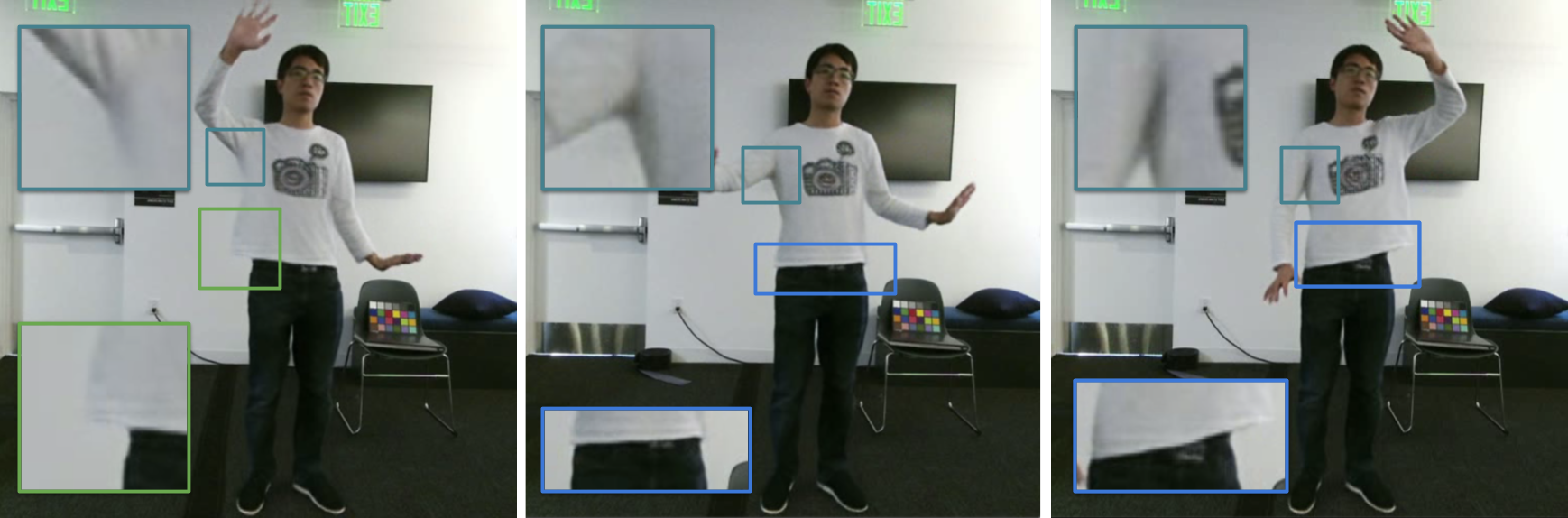}
    \vspace*{-0.7cm}
    %\framebox(350,150){}
    \caption{Images generated by ANR on a challenging animation scenario with clothing deformations. We successfully synthesize images of such deformations and regions outside of the body mesh. As highlighted in the figure, region coverage as well as shading are pose dependent. The example frames are unseen poses for this identity and rendering model. Additionally, the model is able to temporally interpolate between this pose and others and adjust the simulated clothing accordingly. We refer to the supplementary video for a demonstration of temporal stability.
    }
    \label{fig:dynamic_cloth_animation}
\end{figure*}

\noindent Articulated Neural Rendering (ANR) can generate highly detailed representations of articulated objects. Unlike traditional rendering pipelines which use a high resolution mesh and detailed RGB texture for this purpose, we use a low resolution mesh but a high-\emph{dimensional} neural texture to render its detailed RGB appearance from novel views using a neural network. Fig.~\ref{fig:framework} shows an outline of the proposed framework. In the following, we first present an overview of DNR~\cite{thies2019deferred} before presenting our novel ANR framework and its training scheme.

\subsection{Preliminary: Deferred Neural Rendering}\label{sec:dnr}
DNR~\cite{thies2019deferred} translates high dimensional neural latent textures on traditional meshes into RGB images with a neural translator network. Concretely, let $\mathcal{T}$ be a high-dimensional neural texture (a tensor of shape $W\times H\times C$) and let $\mathcal{R}$ be the neural rendering model converting a neural image $\mathbf{I}^{\mathrm{uv}}$ to RGB color. DNR optimizes
\begin{equation}
	\mathcal{T}^*, \mathcal{R}^* =
	\underset{\mathcal{T}, \mathcal{R}}{\mathrm{argmin}} \sum 
	||\mathbf{I}- \mathcal{R}(\mathcal{T}, \mathbf{I}^{\mathrm{uv}})||
\end{equation}
% I remove the k subscript which is conflicting with eq 9 and 10
on all training images $\mathbf{I}$ of the same object. The neural image $\mathbf{I}^{\mathrm{uv}}$ is the result of rasterizing the mesh to image space with the appropriate camera parameters and configuration and texturing it with the neural texture $\mathcal{T}$. The model is fully defined with the optimized texture $\mathcal{T}^*$ and the optimized neural rendering model $\mathcal{R}^*$. DNR uses a U-Net architecture~\cite{ronneberger2015u} for implementing $\mathcal{R}$ and a standard gradient descent based optimization with the ADAM optimizer~\cite{kingma2014adam}.

\subsection{Articulated Neural Rendering}
While DNR is conceptually powerful, it requires accurate 3D geometry to learn view dependent appearance information. Such an assumption is difficult to make in practice, especially for articulated, clothed human appearance, whose shape is often represented by a coarse statistical body shape model~\cite{loper2015smpl} (see Fig.~\ref{fig:dynamic_cloth_animation}). We address this problem in the rendering pipeline, while retaining the ability to work with a coarse, animated mesh to (1)~maintain the high rendering speed and (2)~be able to optimize the final appearance generation in the neural network. Consequently, we re-visit the neural rendering component, $\mathcal{R}$.

Our first observation is that $\mathcal{R}$ not only has to paint the texture inside regions of the neural image $\mathbf{I}^{\mathrm{uv}}$ but also beyond its boundaries due to the use of the coarse mesh. The network should also be aware of the extent to which it needs to paint outside the boundaries of the rasterized mesh. We address both problems by adding a second prediction: an extra single-channel soft mask $\mathbf{M} \in [0, 1]$. The predicted mask is used to blend the generated avatar with the ground-truth background image for training. To prevent the model from predicting a degenerated zero mask (which would minimize the loss to zero), we provide supervision for the mask from an automatic image matting method~\cite{gong2018instance}. Note that while training on the pre-segmented image is another option, this approach is sensitive to erroneous segmentation which prevents the generated images to grow beyond their input coarse body boundaries. Comparing the blended image with the ground truth image allows gradients to flow to the mask which in turn makes it potentially better than the supervised input mask.  

While this addresses the immediate problem of generating content outside of the true geometry silhouette, it leaves geometric details and pose-dependent rendering untouched. We notice that naively increasing the  capacity of the U-Net does not improve generation quality (see Tab. \ref{table:ablation}). Furthermore, we observe that the model cannot consistently render local geometry---a problem that increasingly emerges in the articulated setting when geometry is animated. We address problems with the geometry details and the pose-dependent effects at the same time by splitting the neural rendering network in two stages: $\mathcal{R}_1$ and $\mathcal{R}_2$. Both components are shallow U-Nets and produce renderings at original image resolution. We can inject the normal information into the rendering process by concatenating the rendered normal image and output of $\mathcal{R}_1$ to the input of $\mathcal{R}_2$.  We enforce an additional RGB loss on the first three output channels of $\mathcal{R}_1$ to aid in convergence. The ANR model is defined as
\begin{align}
	\hat{\mathbf{M}}, \hat{\mathbf{I}}, \hat{\mathbf{J}} = \mathcal{R}_2(\mathcal{R}_1(\mathcal{T}, \mathbf{I}^{\mathrm{uv}}), \mathbf{I}^{\mathrm{norm}}),
\end{align}
where $\hat{\mathbf{J}}$ are the first three channels from the result of $\mathcal{R}_1$ and $I^{norm}$ is the rasterized normal image. This model has the necessary capacity and the necessary outputs for handling the articulated neural rendering problem.

\subsection{Loss Functions and Regularization Scheme}

With the higher requirements for stability and level of detail and deformations in the articulated setting, we find that using a simple $\ell_1$ loss is insufficient (see Tab.~\ref{table:ablation}). Furthermore, we observe that it deteriorates performance as the training progresses: once the model learns to reproduce the rough appearance, inaccuracies in tracking and alignment of the mesh have an increasingly negative impact (see Fig.~\ref{fig:ablation}). We use adversarial learning and feature loss computation to guide the model to generate realistic and accurate appearance without having to rely on accurate registration. Our loss function is a weighted sum of the photometric loss $\mathcal{L}_{p}$, feature loss $\mathcal{L}_{feat}$, mask loss $\mathcal{L}_{mask}$, adversarial loss $\mathcal{L}_{adv}$, and total variation loss $\mathcal{L}_{tv_i}$. 
% Importantly, while we use $\mathcal{L}_{p}$ for the intermediate RGB image with a constant weight, we significantly relax its weight for the final image as the training progresses~\cite{huang2020adversarial}. Intuitively, the intermediate image should be constrained to capture the \textit{static} appearance while the later image should have more freedom to create realistic pose-dependent \textit{dynamic} appearance. 
Note that while the rasterization is non-differentiable, ANR is fully differentiable given the precomputed rasterized UV lookup to paint $I^{\mathrm{uv}}$ from the neural texture $\mathcal{T}$.

\boldstart{Pixel Loss:} We enforce an $\ell_1$ loss between the generated RGB and ground truth images as
\begin{equation}
	\mathcal{L}_{p}(\hat{\mathbf{M}}, \hat{\mathbf{I}}, \hat{\mathbf{J}}; \mathbf{M}, \mathbf{I}) = \hat{\mathbf{M}}||\hat{\mathbf{J}} - \mathbf{I}|| + \hat{\mathbf{M}} ||\hat{\mathbf{I}} - \mathbf{I}||,
\end{equation}
% \begin{equation}
% 	\mathcal{L}_{p}(\hat{\mathbf{M}}, \hat{\mathbf{I}}, \hat{\mathbf{J}}; \mathbf{M}, \mathbf{I}) = \hat{\mathbf{M}}||\hat{\mathbf{J}} - \mathbf{I}|| + \alpha \hat{\mathbf{M}} ||\hat{\mathbf{I}} - \mathbf{I}||,
% \end{equation}
% where $\hat{\mathbf{J}}$ are the first three channels of the result from $\mathcal{R}_1$, and  $\alpha = 1-\frac{1}{1+\exp{(-n/k)}}$ is a decay rate dependent on number of epochs and a scaling constant, $n$ and $k$ respectively.\\
where $\hat{\mathbf{J}}$ are the first three channels of the result from $\mathcal{R}_1$.\\
\boldstart{Mask Loss:} Similarly, we use a Binary Cross Entropy loss for the mask
\begin{equation}
	\mathcal{L}_{mask}(\hat{\mathbf{M}}; \mathbf{M}) =  \mathrm{BCE}(\hat{\mathbf{M}}, \mathbf{M})
\end{equation}
For all following loss definitions, we introduce the shorthand $\hat{\mathbf{I}}'$ for the blended version of the generated output with the scene background $\mathbf{B}$ given the predicted mask: $\hat{\mathbf{I}}'=\hat{\mathbf{M}}\hat{\mathbf{I}} + (1-\hat{\mathbf{M}})\mathbf{B}$. \\
\boldstart{Feature Loss:} To increase sharpness in the rendered outputs, we enforce a feature loss~\cite{johnson2016perceptual}:
\begin{equation}
	\mathcal{L}_{feat}(\hat{\mathbf{I}}, \hat{\mathbf{M}}; \mathbf{I}) = \sum_j w_j ||\phi_j(\hat{\mathbf{I}}') - \phi_j(\mathbf{I})||,
\end{equation}
where $\phi_j$ are features from the $j$-th layer of a pretrained feature extractor and $w_j$ is the weight associated with the $j$-th feature loss term.\\
\boldstart{TV Loss:} Since the texture is optimized over multiple frames, slight misalignments can cause the learned texture to have certain high frequency artifacts, especially for small regions such as face and hands. To encourage smooth generated images, we enforce a total-variation loss on both, the mask and the generated image.
\begin{equation}
	\mathcal{L}_{tv}(\hat{\mathbf{I}}, \hat{\mathbf{M}}) = \beta_I TV(\hat{\mathbf{I}}') + \beta_m TV(\hat{\mathbf{M}}),
\end{equation}
where $\beta_I$ and $\beta_m$ are weights associated with Image and mask TV loss respectively (see supp. mat. for a detailed definition of this loss).\\
% \boldstart{Adversarial Loss:} with decaying weight of $\mathcal{L}_p$ in the training we need another loss enforcing realism of the results and encourage the coarse body mask to extend to the true geometry silhouette. 
\boldstart{Adversarial Loss:}  
Adversarial training~\cite{huang2020adversarial} is well-suited for enforcing realism of the results and encourage the coarse body mask to extend to the true geometry silhouette. To encourage a high level of detail in the results, we use a multiscale discriminator $\mathcal{D}$~\cite{wang2018high} and express the loss as 
\begin{equation}
	\mathcal{L}_{adv}(\hat{\mathbf{I}}) = \mathcal{D}(\hat{\mathbf{I}}', 1).
\end{equation}

\boldstart{Total loss:} The loss used to train $\mathcal{R}$ is then given as
\begin{equation}
	\mathcal{L}_{total} = \sum_{i \in \mathfrak{L}} \lambda_i \mathcal{L}_i
\end{equation}
where $\mathfrak{L}=\{p,feat,mask,adv,tv\}$ is the set of all losses.

\subsection{Optimization}

% Despite the extended set of losses, weight balancing and loss decay, we find that for clothing with large surface deformations, the model starts averaging fine textures in areas of high deformation. To mitigate this problem, we propose a \emph{split optimization} strategy. 
Despite the extended set of losses and weight balancing, we find that for clothing with large surface deformations, the model starts averaging fine textures in areas of high deformation. To mitigate this problem, we propose a \emph{split optimization} strategy. 
Specifically, we use a small set of keyframes $\{\mathcal{K}_i\}_{i=1}^n$, capturing \textit{static} salient appearances in the video, to learn the neural texture $\mathcal{T}$ and use the other frames to \textit{dynamically} blend between the appearances in the keyframes in the neural renderer $\mathcal{R}$. 

We select keyframes by greedily adding a small number of frames in the video sequence such that their cumulative silhouette coverage is maximized. This ensures that the entire pose-space is adequately covered to capture texture details at all locations on the body. Using a smaller number of frames (less than 10\% of training frames) reduces the texture averaging. During training, we alternate between training the identity specific neural texture from the keyframes and the rendering network from the remaining frames. Empirically, we observe that this optimization scheme helps the translator network converge up to 5X faster and produces quantitatively better avatars (see Tab.~\ref{table:quantitative}). Overall, our optimization alternates between the following two objectives

\begin{eqnarray}
	\underset{\mathcal{R}}{\mathrm{argmin}} \sum_{}
	\mathcal{L}_{total}(\mathbf{I}, \mathbf{M}, \mathcal{R}(\mathcal{T}, \mathbf{I}^{\mathrm{uv}}, \mathbf{I}^{\mathrm{norm}})), \label{eq9}\\
	\underset{\mathcal{T}, \mathcal{R}}{\mathrm{argmin}} \sum_{k\in \mathcal{K}}
	\mathcal{L}_{total}(\mathbf{I}_k, \mathbf{M}_k, \mathcal{R}(\mathcal{T}, \mathbf{I}_k^{\mathrm{uv}}, \mathbf{I}_k^{\mathrm{norm}})). \label{eq10}
\end{eqnarray}
Note that while Eq.~\ref{eq9} is optimized for all the images, Eq.~\ref{eq10} is applied only to the keyframes to mitigate the geometric misalignment of the coarse body mesh.

\begin{figure*}
    \centering
    \includegraphics[width=\linewidth]{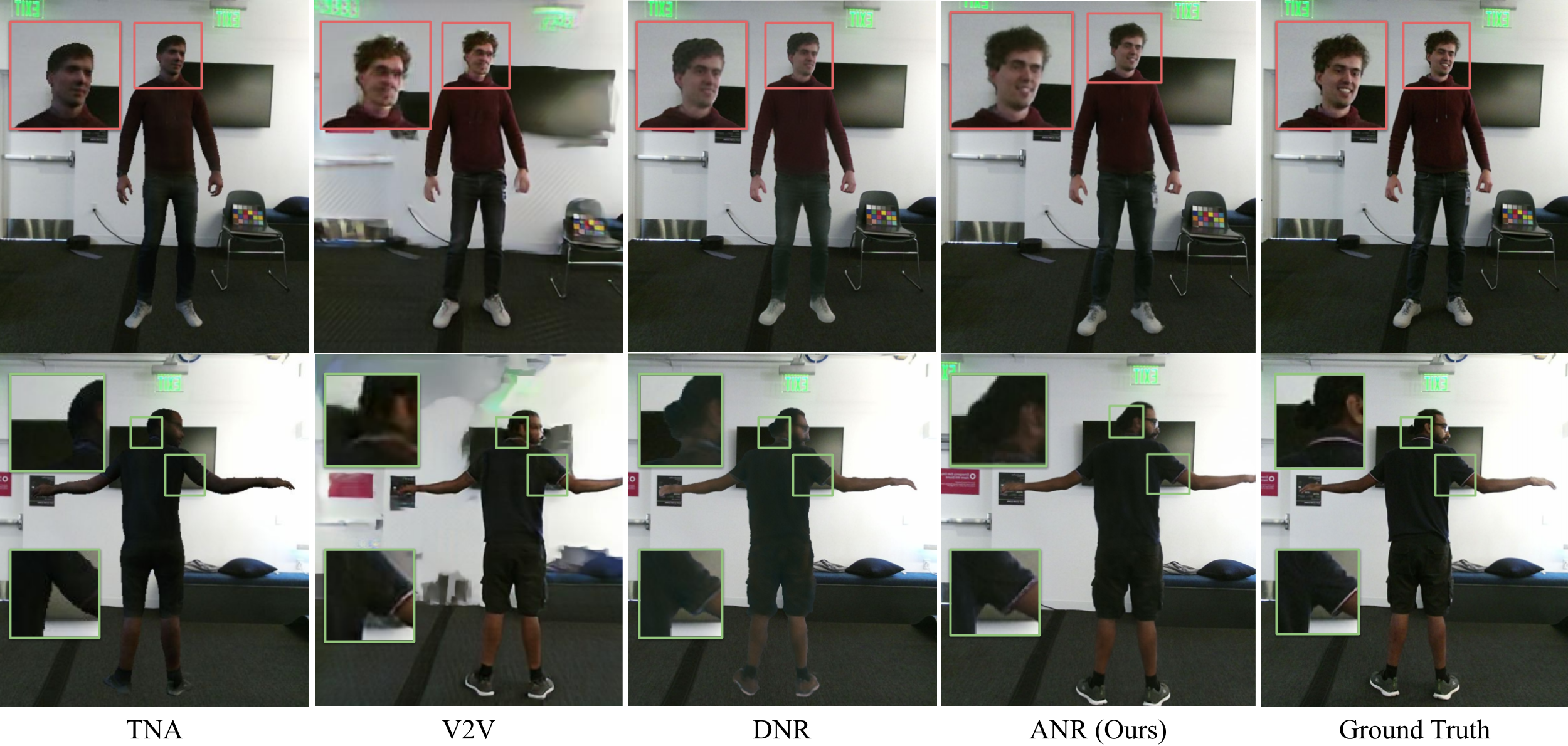}
    \vspace*{-0.7cm}
    \caption{Comparisons for novel pose and view synthesis with Textured Neural Avatar (TNA)~\cite{shysheya2019textured}, vid2vid (V2V)~\cite{wang2018vid2vid} and Deferred Neural Renderer (DNR)~\cite{thies2019deferred}. Our method (ANR) preserves the facial details better compared to competing methods. Additionally, our method is able to capture view dependent structures like hairline and clothing overhang more accurately and leads to more realistic and believable shading.}
    \label{fig:comparison}
    \vspace{-.5cm}
\end{figure*}

\boldstart{Multi-instance Training.} We further extend the training scheme beyond a single capture instance. Since we use the same statistical mesh regardless of identity, allowing us to capture identity information only in the neural texture, our framework can naturally train on multiple identities simultaneously in a single network. During optimization, we select an identity for every step at random and use an identity-specific neural texture $\mathcal{T}_i$ for the respective identity for the update step. The multi-instance training offers the additional benefit that the neural rendering component $\mathcal{R}$ generalizes beyond a single identity and can be used to render new identities by only using a novel neural texture $\mathcal{T}$.

\boldstart{Regularization.} To improve generalization, we additionally employ two training regularization schemes. First, we use the same initialization of $\mathcal{T}$ for all identities by uniformly sampling in $[-1,1]$. Since $\mathcal{R}$ has much larger capacity than $\mathcal{T}$, this strategy prevents the model from using the distinct noise patterns in each randomly initialized $\mathcal{T}$ to memorize the identity and thus encourages decoupling of $\mathcal{T}$ and $\mathcal{R}$. Second, we perturb the input sampling grid with a uniform samples from $[-0.02,0.02]$ and clamp the resulting grid back to $[-1,1]$. This form of data augmentation prevents the network from relying strictly on the spatial extent of the sampling grid as the ground truth human silhouette can exist outside the rasterized coarse body model.

\section{Experiments}

We use the ANR pipeline to build a realistic virtual human avatar pipeline: we assume a setting where a user performs a recording of themselves with accurate tracking, in which his/her full appearance is visible, to create an avatar model. To ease the tracking, we capture 6 videos using a Kinect V2 where the depth data is only used for tracking. Each video is about 3$\sim$5 min long. We obtain a coarse mesh in real-time by solving an inverse kinematic problem to fit the posed body shape to the 3D point cloud similar to~\cite{walsman2018articulated}, making use of additional detected body keypoints~\cite{cao2018openpose}. Our dataset is harder than the previously released iPer dataset \cite{si2018multistage} as our actors are not centered and are free to move anywhere in the frame. As parametric body model, we use a blendshape-based, SMPL-like~\cite{loper2015smpl} human model to provide the coarse mesh structure. The model is coarse and has only 1831 vertices and 3658 faces; the skeletal rig has 74 joints.

\subsection{Implementation Details}

We use a variant of Pix2Pix~\cite{isola2017image,wang2018high} for both $\mathcal{R}_1$ and $\mathcal{R}_2$ and train the model on $1024 \times 1024$ image resolution. The images are normalized to the range [-1,1]. Each identity is encoded in a $256 \times 256 \times 8$ neural texture. 
For each recorded sequence, we use the first 1500 frames to train $\mathcal{R}$ and about 150 key frames to train $\mathcal{T}$. The remaining images are used as test set. We augment the data with random cropping and random rescaling by a factor $f \sim [0.5,1.25]$.

\subsection{Evaluation}
\begin{table}
\begin{center}
\caption{Results of novel pose synthesis of avatars learned using different methods. Our model is trained on all identities simultaneously.}
%\vspace*{-0.3cm}
\label{table:quantitative}
\resizebox{0.48\textwidth}{!}{ 
\begin{tabular}{l|lllll|l}
\hline\noalign{\smallskip}
 & SSIM $\uparrow$ & ~FLIP $\downarrow$ & ~ LPIPS $\downarrow$ & ~rIPFIP ~$\uparrow$ & mFID $\downarrow$ & User Study \\
\noalign{\smallskip}
\hline
\noalign{\smallskip}
V2V & 0.9252 & ~0.0363 & ~0.0703~ & ~ - & 140 & 8\%\\
TNA & 0.9366 & ~\underline{0.0323}~ & ~0.1198 & ~-2.6\% & 150  & 3\% \\
DNR & \underline{0.9398} & ~0.0342 & ~0.0918 & ~\underline{7.7\%} & 92 & 9\%\\
ANR & \textbf{0.9738} & \textbf{0.0289} & \textbf{0.0508} & \textbf{18.6\%} &74 &\textbf{81.6\%}  \\
\hline
\end{tabular}
}
\end{center}
\vspace*{-0.5cm}
\end{table}

\label{sec:eval}
\boldstart{Baseline and Metric}: We include a comparison with two baselines: Textured Neural Avatar (TNA)~\cite{shysheya2019textured} and vid2vid (V2V)~\cite{wang2018vid2vid}. These methods are fundamentally different and span the space of 2D (V2V) and 3D (TNA) inference approaches, whereas we aim to find a middle ground. We also present comparisons to baseline DNR~\cite{thies2019deferred}, trained with additional feature losses for a fairer comparison. Fig.~\ref{fig:comparison} shows these comparisons. Evidently, ANR preserves the facial details better compared to competing methods.  Additionally, it is able to capture view dependent structures like hairline and clothing overhang more accurately and leads to more realistic and believable shading. We also quantify these renderings using the standard SSIM, LPIPS, FLIP \emph{supervised metrics} on held out test set frames, and the mFID \emph{unsupervised metric} on human-figure-only avatars in novel poses. Tab.~\ref{table:quantitative} shows these comparisons. Our model outperforms the competing approaches on these benchmarks by a notable margin on all metrics. 

\begin{figure}
    \centering
    \includegraphics[width=\linewidth]{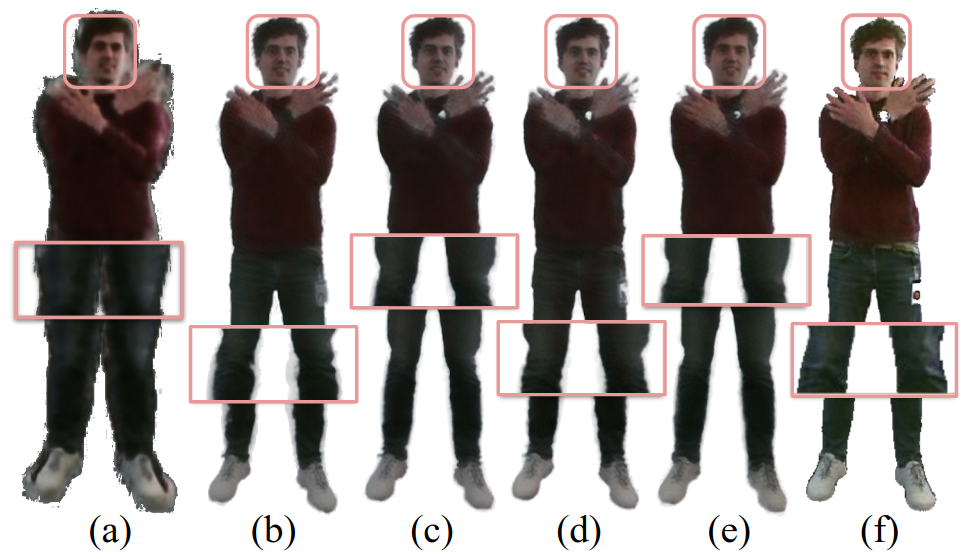}
    \vspace*{-0.7cm}
    \caption{Qualitative ablation study for each term in the loss function and the normal channels on a multi-identity model. The first four columns do not use normal information: \textbf{(a)} Only pixel loss; \textbf{(b)} Pixel and feature losses; \textbf{(c)} Pixel, feature, and mask losses; \textbf{(d)} Pixel, feature, mask, and TV losses; \textbf{(e)} All losses + normal; \textbf{(f)} All loses + normal + split optimization. Notice that the local textures and facial details are better preserved with split optimization.}
    \label{fig:ablation}
\end{figure}

\begin{figure*}
    \centering
    \includegraphics[width=\linewidth]{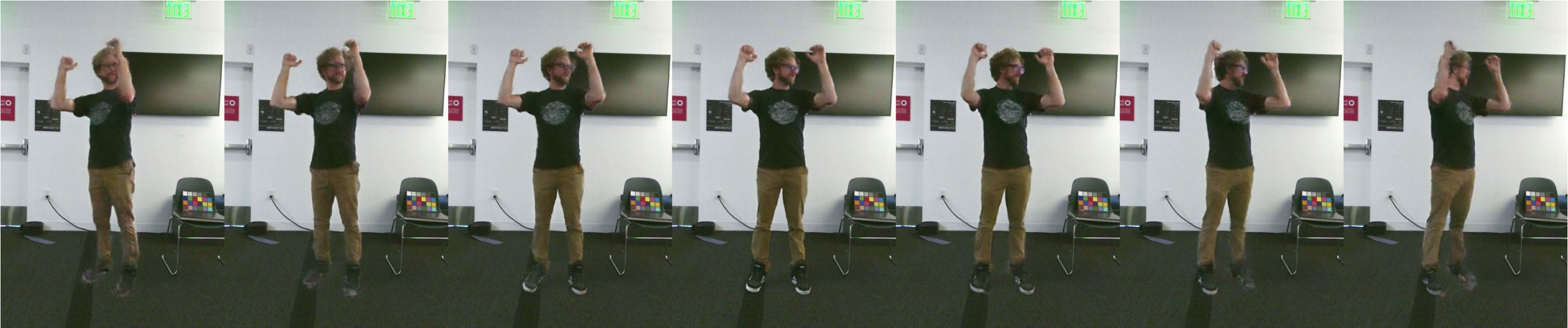}
    %\framebox(35h0,150){}
    \vspace*{-0.6cm}
    \caption{Viewpoint generalization demonstration. The proposed model model is robust to viewpoint variation, even for unseen poses, and shows high level of detail.}
    \label{fig:novel_view}
    \vspace*{-0.5cm}
\end{figure*}

\begin{figure}
    \centering
    \includegraphics[width=\linewidth]{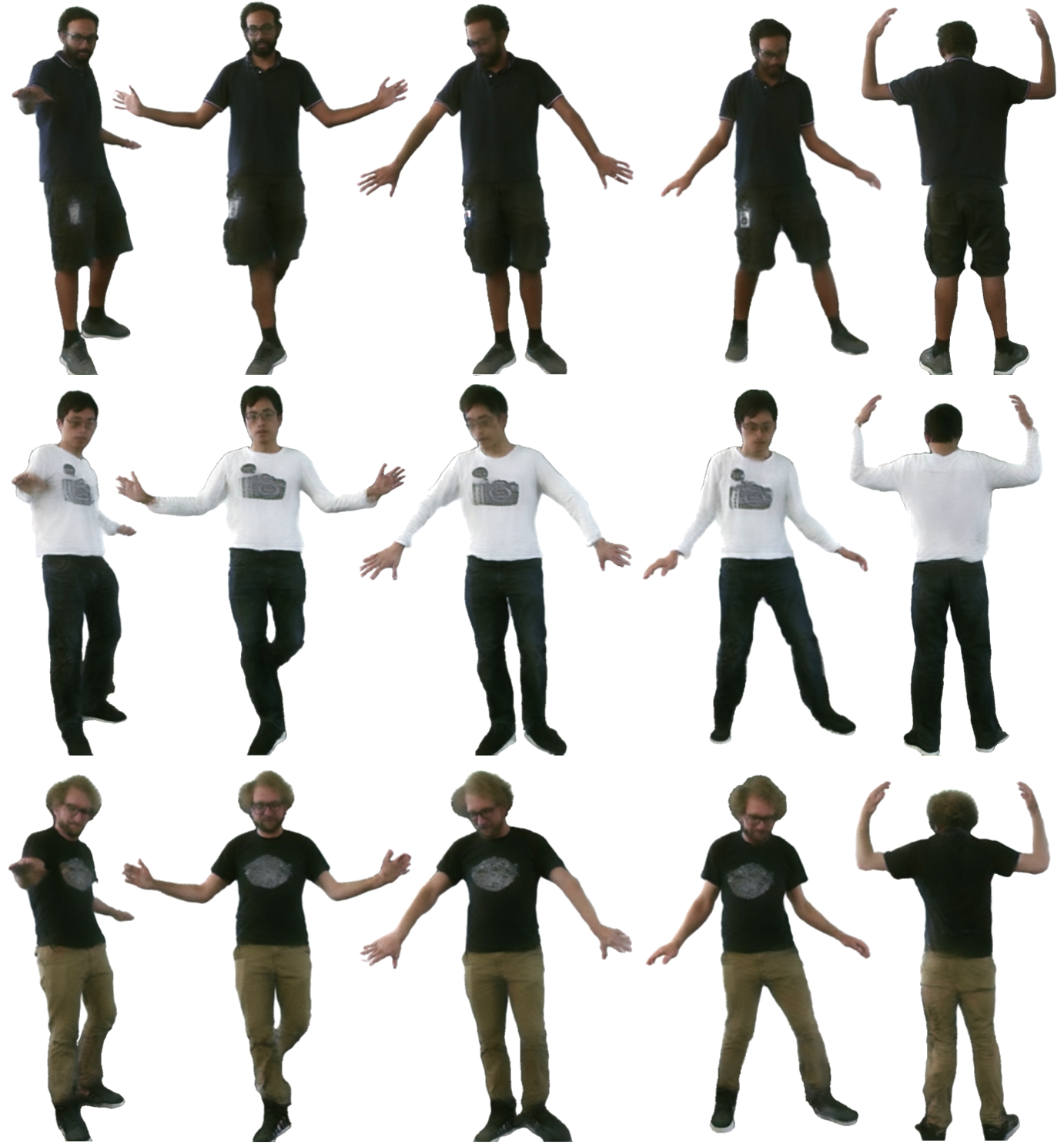}
    \vspace*{-0.7cm}
    \caption{Avatar animation example. Any motion capture data that can be used to animate the base mesh can be used to drive the avatar. All avatars shown here are rendered using a single neural network.}
    \label{fig:retargeting}
    \vspace{-.6cm}
\end{figure}

\begin{figure}
    \centering
    \includegraphics[width=.9\linewidth]{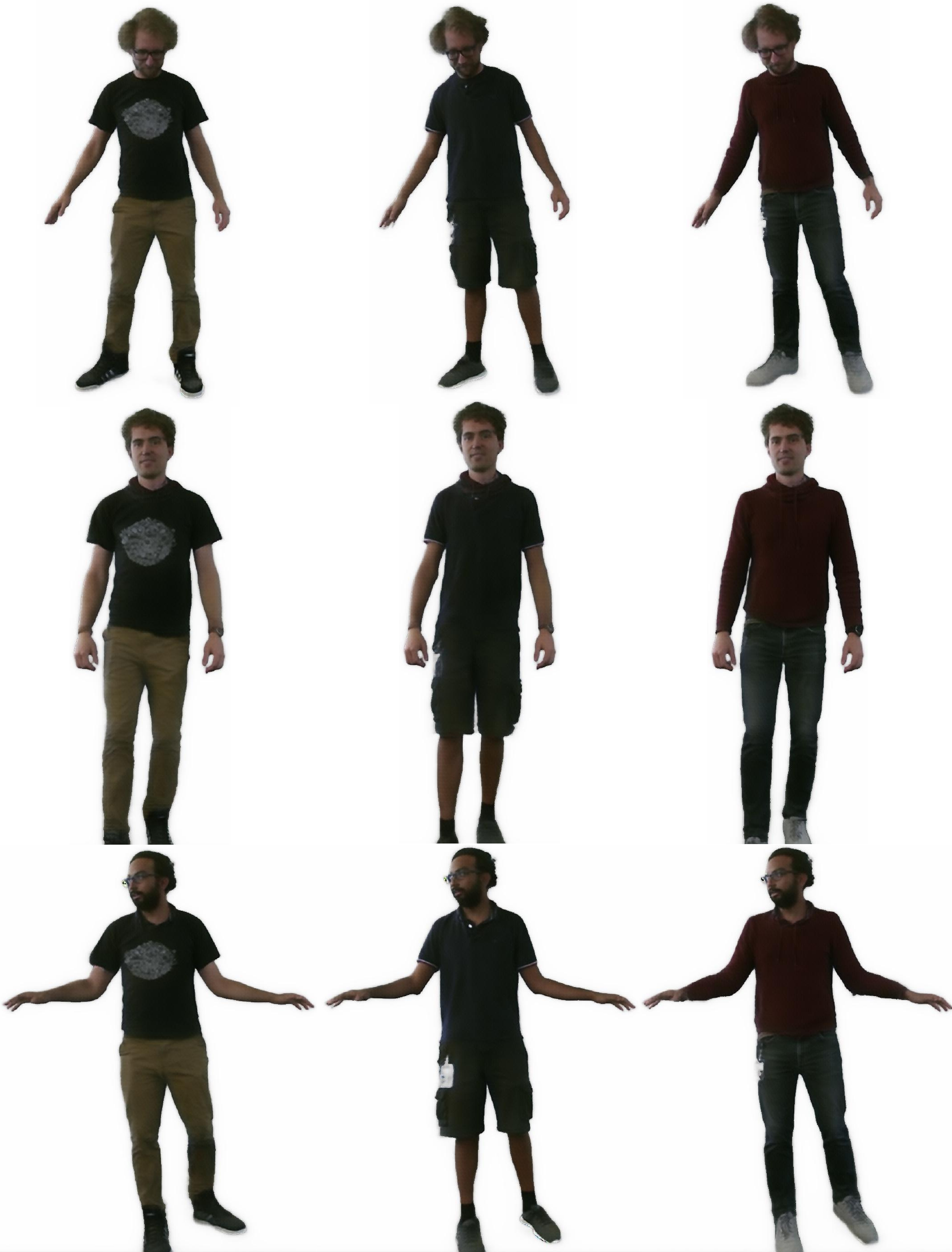}
    \vspace*{-0.3cm}
    \caption{Virtual Try-On example. Our method enables texture mixing by swapping the regions of the neural texture of the desired RGB appearance. This example validates the disentanglement of appearance and neural shading network when ANR is trained on multiple identities.}
    \label{fig:texturemixing}
    \vspace{-.6cm}
\end{figure}

\setlength{\tabcolsep}{1.4pt}
\begin{table}
\begin{center}
\caption{Loss and model ablation study for the ANR model. The model ablations marked with (-so) are run without the suggested split optimization strategy.}
\vspace*{-0.3cm}
\label{table:ablation}
\resizebox{0.32\textwidth}{!}{ 
\begin{tabular}{l|llll}
\hline\noalign{\smallskip}
& SSIM $\uparrow$~ & LPIPS $\downarrow$~ & FLIP $\downarrow$~ \\
\hline
\multicolumn{4}{c}{\emph{Loss ablation~}} \\
Pixel only & 0.968 & ~0.086~ & ~0.029~ \\
Pixel+feat & 0.966 & ~0.065~ & ~0.033~ \\
Pixel+feat+TV  & 0.963 & ~0.064~ & ~0.032~ \\
\hline
\multicolumn{4}{c}{\emph{Model ablations~}} \\

1 stage(-so) & 0.962 & ~0.070~ & ~0.036~ \\
1 stage  & 0.965 & ~0.063~ & ~0.034~ \\
2 stage(-so) & 0.968 & ~0.058~ & ~0.032~ \\
\hline
Ours & \textbf{0.974} & \textbf{0.050} & \textbf{0.028}
\end{tabular}
}
\end{center}
\vspace*{-0.5cm}
\end{table}
\setlength{\tabcolsep}{2pt}

\boldstart{Ablation Study:} To quantify the effectiveness of the proposed improvements, we run two ablation studies. % and present the results in Fig.~\ref{fig:ablation} and Tab.~\ref{table:ablation}.
Fig.~\ref{fig:ablation} shows the rendered results with different loss terms removed. We notice that without the mask and feature loss (Pixel only), the model produces unrealistically ``fat'' or ``thin'' avatars. The feature loss (Pixel+feat) improves the visual quality. Adding the normals improves the level of detail in the reconstruction and aids in reasoning about self occlusion and temporal consistency (shown in the supplementary video), split optimization drastically improves the level of detail. Note the jump in perceptual quality of the rendered face using the split-optimization scheme. Lastly, we show that our two-stage neural render with intermediate normal injection outperforms the single-stage approach with the same capacity, validating our network design choice. This trend is quantitatively confirmed in Tab.~\ref{table:ablation}.

\boldstart{User Study:} While SSIM, LPIPS, or FLIP are the most widely-used metrics for generative tasks, they are merely proxy metrics which do not pay attention to salient regions (\eg, for faces or patterns on shirts) and do not strictly measure perceptual quality.
To demonstrate the efficacy of our method, we conduct a 4-alternative forced-choice perceptual study with 80 participants, where users were given a choice to pick the best avatar out of the results generated from TNA, Vid2Vid, DNR, and our ANR. Each person was presented with 20 stimuli of avatars in novel poses for 5s (see supplementary material for details). ANR was preferred $\textbf{81.6\%}$ of the time. Furthermore, to test the photorealism of our avatars, we conducted another 2-Alternative forced choice study where users were presented a real image and an image of our avatar in  different poses, and asked to pick the real image. Our model was able to fool users \textbf{30\%} of the time (50\% being random chance) in this test. This shows the realistic rendering performance of our model.

\boldstart{Model Efficiency:} We calculate the relative improvement in LPIPS of each approach~(x) over vid2vid~(V2V) scaled by factor of improvement in number  of parameters (\#p) as
\[
\mathrm{rIPFIP}(x)=\frac{\mathrm{LPIPS}_{v2v}-\mathrm{LPIPS}_{x}}{\mathrm{LPIPS}_{v2v}}*\frac{log(\#p_{v2v}/\#p_{x})}{log(\#p_{V2V})}
\]
Particularly, this metric lies in (-$\infty$,1] and reaches a maximum value for ground truth images.
This metrics highlights that we benefit from our \emph{design choices} compared to DNR, and not solely from differences in capacity.

\section{Applications}

We use a single ANR model to digitize and render avatars for several applications. Please refer to the supplementary video for more examples.

\boldstart{Novel View Synthesis:} To render the avatar from novel views, we only need to rasterize the tracked mesh using the scene camera parameters to create the UV lookups. The avatar can be readily generated using the neural renderer $\mathcal{R}$. See Fig.~\ref{fig:novel_view} for an illustration and the supplementary video for additional results.  The viewpoint stability is unlike most image-based CNN approaches, which often synthesize inconsistent appearance with varying viewpoints~\cite{wang2018vid2vid}.

\boldstart{Animation:} The learned neural identity can be retargeted to any motion from a motion capture database. Fig.~\ref{fig:retargeting} shows renderings of the same motion sequence from multiple views. Importantly, our model adds vivid and realistic pose-dependent deformation to the rendered avatar, which is not possible for other methods using skinned, but coarse meshes~\cite{huang2020arch}. Fig.~\ref{fig:dynamic_cloth_animation} provides a detailed view of the pose-dependent deformation appearance generation.

\boldstart{Replacement of Textures / Virtual Try-On:} The learned neural texture is not directly interpretable. However, for two identities trained on the same neural rendering network, we can swap parts of the neural volume to generate identities with swapped faces/clothing items, as shown in Fig.~\ref{fig:texturemixing}. This is unlike fully 3D based approaches which require detailed captures for each new avatar~\cite{guo2019relightables}. 

\section{Conclusion}
We introduce Articulated Neural Rendering (ANR), a novel neural rendering framework, to generate high-quality virtual avatars with arbitrary skeletal animations and viewpoints. Key to our work is the ability to account for geometric misalignment and pose-dependent surface deformation. Our solutions are carefully integrated into an end-to-end learning framework with a novel neural rendering architecture and adjusted optimization scheme. Additionally, ANR can render multiple avatars using a single neural rendering model. Through decoupling of texture and geometry it allows for mixing and editing of appearance. For future work, we see potential directions in further mitigating the impact of geometric misalignment and improving resiliency towards large pose tracking errors as well as incorporating environmental lighting into the rendering process.

\boldstart{Acknowledgement} This work was done while AR, JT, and CS were at Facebook. We thank Tiancheng Zhi and Tony Tung for help with data processing and Michael Zollh{\"{o}}efer for the fruitful discussions. 
{\small
\bibliographystyle{ieee_fullname}
\bibliography{egbib}
}

\end{document}